\documentclass[sigconf]{acmart}

\AtBeginDocument{%
  \providecommand\BibTeX{{%
    \normalfont B\kern-0.5em{\scshape i\kern-0.25em b}\kern-0.8em\TeX}}}

\setcopyright{acmcopyright}
\copyrightyear{2022}
\acmYear{2022}
\setcopyright{acmcopyright}\acmConference[ASE '22]{37th IEEE/ACM International Conference on Automated Software Engineering}{October 10--14, 2022}{Rochester, MI, USA}
\acmBooktitle{37th IEEE/ACM International Conference on Automated Software Engineering (ASE '22), October 10--14, 2022, Rochester, MI, USA}
\acmPrice{15.00}
\acmDOI{10.1145/3551349.3556900}
\acmISBN{978-1-4503-9475-8/22/10}

\usepackage{mathtools}
\usepackage{pythonhighlight}
\usepackage{algpseudocode}
\usepackage{algorithm}
\usepackage{pifont}
\usepackage{xcolor}
\usepackage{subcaption}
\usepackage{hyperref}

\newcommand{\cmark}{\ding{51}}%
\newcommand{\xmark}{\ding{55}}%

\newcommand\encircle[1]{%
	\tikz[baseline=(X.base)] 
	\node (X) [draw, shape=circle, inner sep=0.75, fill=white!80!red, text=black, draw=white!40!red] {\bf\footnotesize\sffamily #1};%
}

\usepackage{amsmath}
\usepackage{amsthm}
\usepackage{mathtools, nccmath}
\usepackage{wrapfig}
\usepackage{comment}
\usepackage{framed}
\usepackage{graphicx}
\usepackage{tikz}
\usetikzlibrary{backgrounds}
\usetikzlibrary{arrows,shapes}
\usetikzlibrary{tikzmark}
\usetikzlibrary{calc}
\usepackage[edges]{forest}
\usetikzlibrary{arrows.meta}
\colorlet{linecol}{black!75}

\definecolor{dark-red}{RGB}{180,0,0}
\definecolor{dark-green}{RGB}{0,180,0}

\begin{document}

\title{AST-Probe: Recovering abstract syntax trees from hidden representations of pre-trained language models}



\newcommand{\chunk}[2]{%
	\fcolorbox{black}{yellow}{\bfseries\sffamily\scriptsize#1}%
   {$\blacktriangleright$#2$\blacktriangleleft$}%
}
\newcommand{\note}[1]{\chunk{Note}{{\textcolor{blue}{\textsl{#1}}}}}
\newcommand{\martin}[1]{\chunk{Martin}{{\textcolor{teal}{\textsl{#1}}}}}
\newcommand{\houari}[1]{\chunk{Houari}{{\textcolor{red}{\textsl{#1}}}}}
\newcommand{\jose}[1]{\chunk{José}{{\textcolor{purple}{\textsl{#1}}}}}
\newcommand{\jesus}[1]{\chunk{Jesús}{{\textcolor{orange}{\textsl{#1}}}}}

\author{Jos\'e Antonio Hern\'andez L\'opez}
\affiliation{\institution{University of Murcia}\country{Murcia, Spain}}
\email{joseantonio.hernandez6@um.es}
\authornote{Both authors contributed equally to the paper}

\author{Martin Weyssow}
\affiliation{DIRO, \institution{University of Montreal}\country{Montreal, Canada}}
\email{martin.weyssow@umontreal.ca}
\authornotemark[1]

\author{Jes\'us S\'anchez Cuadrado}
\affiliation{\institution{University of Murcia}\country{Murcia, Spain}}
\email{jesusc@um.es}

\author{Houari Sahraoui}
\affiliation{DIRO, \institution{University of Montreal}\country{Montreal, Canada}}
\email{sahraouh@iro.umontreal.ca}

\begin{abstract}
    The objective of pre-trained language models is to learn contextual representations of textual data. Pre-trained language models have become mainstream in natural language processing and code modeling.
    Using probes, a technique to study the linguistic properties of hidden vector spaces, previous works have shown that these pre-trained language models encode simple linguistic properties in their hidden representations. However, none of the previous work assessed whether these models encode the whole grammatical structure of a programming language. In this paper, we prove the existence of a \textit{syntactic subspace}, lying in the hidden representations of pre-trained language models, which contain the syntactic information of the programming language. We show that this subspace can be extracted from the models' representations and define a novel probing method, the AST-Probe, that enables recovering the whole abstract syntax tree (AST) of an input code snippet. In our experimentations, we show that this syntactic subspace exists in five state-of-the-art pre-trained language models. In addition, we highlight that the middle layers of the models are the ones that encode most of the AST information. Finally, we estimate the optimal size of this syntactic subspace and show that its dimension is substantially lower than those of the models' representation spaces. This suggests that pre-trained language models use a small part of their representation spaces to encode syntactic information of the programming languages.
\end{abstract}




\maketitle

\section{Introduction}
\label{sec:intro}
The naturalness hypothesis of software~\cite{hindle2016naturalness} states that code is a form of human communication bearing similar statistical properties to those of natural languages. Since this key finding, much work has focused on the application of deep learning (DL) and natural language processing (NLP) to learn semantically meaningful representations of source code artifacts~\cite{allamanis2018survey}. These representations are vectors that lie in hidden representation spaces of the DL model. These DL-based approaches have drastically improved the automation of a wide range of code-related tasks such as code completion~\cite{svyatkovskiy2021fast, lu2022reacc}, code search~\cite{cambronero2019deep, husain2019codesearchnet} or code summarization~\cite{bui2021self}. In particular, the state-of-the-art leverages pre-trained language model architectures such as those of BERT~\cite{devlin2018bert, liu2019roberta} or GPT~\cite{Radford2018-GPT, radford2019-gpt2, brown2020language-gpt3} to learn representations of source code. Among others, CodeBERT~\cite{feng2020codebert}, GraphCodeBERT~\cite{guo2020graphcodebert}, CodeT5~\cite{wang2021codet5} or Codex~\cite{chen2021evaluating} have shown great performance and flexibility to lots of downstream tasks thanks to task-specific fine-tuning objectives.

A wide range of previous work has shown the great potential of these models in automating many tasks. However, what these models learn in their hidden representation spaces remains an open question. In this paper, we analyze the hidden representation spaces of pre-trained language models to broaden our understanding of these models and their use in automated software engineering.
In the NLP field, \textit{probing} emerged as a technique to assess whether hidden representations of neural networks encode specific linguistic properties~\cite{adi2016fine, belinkov2016probing, hewitt2019structural}. A probe is a simple classifier that takes the model's hidden representations as input and predicts a linguistic property of interest. Recently, some previous works relying on probes have shown that the hidden representation spaces of these models encode linguistic properties of the input code and underlying programming language. Among others, Wan et. al~\cite{wan2022they} showed that unlabeled binary trees can be recovered from the representations of an input code. Even though this property is non-trivial, we foresee that hidden representations of pre-trained language models encode more complex constructions of the programming languages. In particular, previous contributions have not proposed a probing method that can probe the whole grammatical structure, \textit{i.e.}, the abstract syntax tree (AST), of a programming language in the hidden representation spaces of pre-trained language models. In this work, \textit{we aim to bridge this gap and determine whether pre-trained language models encode the whole ASTs of input code snippets in their hidden representation spaces.}

We propose the \textit{AST-Probe}, a novel probing approach that evaluates whether the AST of an input code snippet is encoded in the hidden representations of a given pre-trained language model. Our approach relies on the hypothesis that there exists a \textit{syntactic subspace}, lying in the hidden representation spaces of the pre-trained language model, that encodes the syntactic information of the input programming language related to the AST. Using our probe, we aim to predict the whole AST of an input code and thus show that pre-trained language models implicitly learn the programming language's syntax. More specifically, the idea is to project the hidden representation of an input code to the syntactic subspace and use the geometry of this subspace to predict the code AST. 

We evaluate the AST-Probe on five state-of-the-art pre-trained language models to demonstrate the generalizability of the probe and provide in-depth analysis and discussion. We also attempt to estimate the optimal size of the syntactic subspace, \textit{i.e.}, its compactness, to understand how many dimensions are used by the pre-trained language models to store AST-related syntactic information of the programming languages.

The contributions of this work are the following.
\begin{enumerate}
    \item AST-Probe: A language-agnostic probing method able to recover whole ASTs from hidden representations of pre-trained language models to assess their understanding of programming languages syntax.
    
    \item A representation of ASTs as a compact tuple of vectors adapted from~\cite{shen2018straight} that can be used for probing pre-trained language models.
    
    \item An extensive evaluation of the AST-Probe approach on a variety of pre-trained language models: CodeBERT~\cite{feng2020codebert},  GraphCodeBERT~\cite{guo2020graphcodebert}, CodeT5~\cite{wang2021codet5}, CodeBERTa~\cite{wolf2019huggingface} and  RoBERTa~\cite{liu2019roberta}, and programming languages: Python, Javascript and Go.
    
    \item An estimation of the compactness of the optimal syntactic subspace by comparing different dimensions of possible subspaces.
\end{enumerate}

\noindent\textbf{Organization.} Section~\ref{sec:background} gives a brief overview of the technical background and discusses related work. In Section~\ref{sec:approach}, we describe the AST-Probe in-depth. In Section~\ref{sec:experiments} and ~\ref{sec:results}, we go through our experimental setup and discuss the results of our experiments. Finally, in Section~\ref{sec:discussion}, we discuss the main findings of this work and broader analysis. We close this paper in  Section~\ref{sec:conclusion} with future work and a conclusion.

\section{Background and related work}
\label{sec:background}
\subsection{Pre-trained language models}

Traditional NLP models were designed to be task-specific, \textit{i.e.}, trained on specific data, and using specific learning objective functions. In recent years, this paradigm has shifted to pre-training large language models on large text corpora in a self-supervised fashion. These models are pre-trained using relatively general learning objectives to learn representations of the pre-training data. They can then be fine-tuned on a wide range of task-specific data and generally show impressive results~\cite{min2021recent}.

Modern pre-trained language models are based on the Transformer architecture~\cite{vaswani2017attention}. Pre-trained language models can be roughly classified into three categories~\cite{min2021recent}: \textit{autoregressive language models, masked language models, and encoder-decoder language models}. Autoregressive language models' objective is to learn to predict the next word in a sequence by expanding a history made of the previous words through time steps (\textit{e.g.}, GPT-3~\cite{brown2020language-gpt3}, CodeGPT~\cite{lu2021codexglue} or Codex~\cite{chen2021evaluating}). Masked language models are trained with the objective to predict a masked word given the rest of the sequence as context (\textit{e.g.}, BERT~\cite{devlin2018bert}, CodeBERTa~\cite{wolf2019huggingface}, CodeBERT~\cite{feng2020codebert} and GraphCodeBERT~\cite{guo2020graphcodebert}). Finally, encoder-decoder language models can be trained on a sequence reconstruction or translation objective (\textit{e.g.}, T5~\cite{raffel2019exploring}, BART~\cite{raffel2019exploring}, CodeT5~\cite{wang2021codet5} and PLBART~\cite{ahmad2021unified}). 

A language model takes a sequence of tokens as input and produces a set of contextualized vector representations. More concretely, let us denote $w_0,\dots,w_n$ a sequence of $n+1$ tokens. A language model of $\mathcal{L}$ transformer layers consists of the calculus of the final embeddings in the following way:
$$H^\ell = \text{Transformer}_\ell(H^{\ell-1})\:, \forall\:\ell=1,\dots, \mathcal{L}$$
where $H^\ell=[h^\ell_0,\dots,h^\ell_n]$, and each vector $h^\ell_i$ corresponds to the embedding of the token $w_i$. $\text{Transformer}_\ell$ is a transformer layer made of multi-head attention layers, residual connections, and a feed-forward neural network. $H^0$ refers to the initial embeddings computed as a sum of the positional encodings and the initial token embeddings. In this paper, we are interested in the syntactic information encoded in the vectors $h^\ell_i \in H^\ell$ for $\ell=1,\dots, \mathcal{L}$. 

Finally, even though we focus on hidden vector representations in this work, it is worth mentioning that the attention layers of these transformer architectures contain valuable information. Several previous works have attempted to discover what information is encoded in the attention heads of transformer-based models of code. For instance, Wan et. al~\cite{wan2022they} showed that the attention is aligned with the motif structure of the AST, Sharma et. al~\cite{sharma2022exploratory} explored the attention layers of BERT pre-trained on code, and Paltenghi and Pradel~\cite{paltenghi2021thinking} compared the attention weights with the reasoning of skilled humans. As we specifically aim at probing deep learning models, we focus the next two sections on this technique and its usage in the related work.


\subsection{Probing in NLP}

Pre-trained language models are powerful models that have achieved state-of-the-art results not only on NLP but also in code-related tasks. However, one issue that we aim to tackle in this paper is that it is not possible to get a direct understanding of to what extent these models capture specific properties of a language. In this view, \textit{probing classifiers} are used to probe and analyze neural networks. A probe is a simple~\footnote{The term \textit{simple} usually means minimally parameterised~\cite{maudslay2020tale}.} classifier trained to predict a particular linguistic property from the hidden representations of a neural network~\cite{belinkov2016probing}. If the probe performs well, it is said that there is evidence that the neural network representations embed the linguistic property. For instance, Belinkov et al.~\cite{belinkov2017neural} used this framework to perform part-of-speech tagging using the representations of a neural machine translation model. Another example taken from Adi et al.~\cite{adi2016fine} is trying to predict an input sequence length from its hidden vector representation.

In the context of syntax, syntactic probes are used to assess if word representations encode syntactic information of the input language. One example is the \textit{Structural Probe} proposed by Hewitt and Manning~\cite{hewitt2019structural} that is used to assess if dependency trees are embedded in the hidden representations of language models. The procedure consists of training a linear transformation whose target vector space has the following property: \textit{the squared distance between two transformed word vectors is approximately the distance between these two words in the dependency parse tree}. Once the probe is trained, the dependency tree can be reconstructed using the distance between each pair of words in the linearly-transformed representation space~\cite{hewitt2019structural,maudslay2020tale,white2021non}.


In this paper, we adopt similar reasoning to assess if pre-trained language models of code encode the AST in their hidden representations. Our probe is also syntactic, but the target tree is of a different type. We consider ASTs, which are more complex constructions than dependency trees as they contain labeled non-terminal nodes that are not seen in the sequence of input tokens. Moreover, our probing approach is different as we consider an orthogonal projection rather than an arbitrary linear transformation. It allows us to thoroughly define the syntactic subspace (see Section~\ref{sec:approach}).




\subsection{Probing in pre-trained language models of code}

The idea of probing languages models trained on code has been recently introduced, and thus only a few related works have been produced on this research topic over the past couple of years. 

Karmakar and Robbes~\cite{karmakar2021pre} proposed a first method for probing pre-trained language models of code. Through a set of four probing tasks, the authors analyzed the ability of pre-trained language models to capture some simple properties of programming languages. In their work, they concluded that these models implicitly capture some understanding of code in their hidden representations. For instance, they show that using their probes, the cyclomatic complexity of a program or the length of an input code can be predicted from a model's hidden representations. Subsequently, Troshin and Chirkova~\cite{troshin2022probing} extended their work by incorporating more models and more probing tasks to broaden the analysis. Finally, Wan et. al~\cite{wan2022they} designed a more complex probe that can extract unlabeled binary trees from code representations.

Although previous works have probed interesting linguistic properties, they have not focused on designing a probe that can determine if a pre-trained language model captures the full grammatical structure of programming languages. That is, previous works' probing methods do not allow to recover \textit{whole ASTs} (including both terminal and non-terminal nodes). In particular, Troshin and Chirkova~\cite{troshin2022probing} assessed whether pre-trained models understand syntax by predicting several properties extracted from the AST such as token depths and paths. Wan et. al~\cite{wan2022they}'s probe is able to predict unlabeled binary trees at most, which are simplified versions of ASTs that do not encode non-terminal nodes.

In this work, we propose a probing method that allows recovering whole ASTs from pre-trained language models' hidden representations. The key idea of the probe is the assumption of the existence of a syntactic subspace in the hidden representation spaces of these models that encodes the programming languages' syntax. This assumption enables us to estimate the dimension of the subspace and draw conclusions about how compact this information is stored in the pre-trained language models. Thus, our method entails an advance in the state-of-the-art since it substantially improves the latter by assessing the models' understanding of the whole grammatical structure of programming languages.

\section{The AST-Probe approach}
\label{sec:approach}

In this paper, we propose AST-Probe, a probe to determine if pre-trained language models implicitly learn the syntax of programming languages in their hidden representation spaces. To this end, we assess if ASTs can be reconstructed using the information from these hidden representation spaces that the models learn from sequences of code tokens.

The AST-Probe looks for a syntactic subspace~\footnote{We name this vector space the \textit{syntactic subspace}, a term borrowed from~\cite{chi2020finding} that best describes the targeted subspace.} $\mathcal{S}$ in one of the hidden representation spaces of a model that is likely to contain most of the syntactic information of the language. We obtain the subspace $\mathcal{S}$ by learning an orthogonal projection of the representations of input code sequences to this subspace. Then, using the geometrical properties of $\mathcal{S}$, we show how it is possible to reconstruct the whole AST of an input code snippet. The AST-Probe must operate on real vectors to make predictions. Therefore, we first reduce the ASTs of input codes to compact tuples of vectors. The whole process is summarized in the following graph:
$$
\text{AST}\:\overset{\S\ref{ast2binary}}{{\longleftrightarrow}}\:\text{binary tree } \overset{\S\ref{binary2tuple}}{{\longleftrightarrow}}\:(\pmb{d}, \pmb{c}, \pmb{u})\:\overset{\S\ref{probe}}{{\longleftarrow}}\:\text{AST-Probe}
$$
Given a code snippet, we extract its AST and convert it into a tuple of vectors $(\pmb{d}, \pmb{c}, \pmb{u})$ that compresses all the AST information. To this end, we adapt the approach proposed by Shen et al~\cite{shen2018straight} for natural language constituency parsing. The conversion is performed in two steps: (1) we transform the AST into a binary tree (Sect.~\ref{ast2binary}) and (2) we extract the tuple from the binary tree (Sect.~\ref{binary2tuple}). Note that these two transformations are bidirectional which allows us to recover the binary tree and the AST from the tuple of vectors. Finally, our AST-Probe aims at predicting the vectors $(\pmb{d}, \pmb{c}, \pmb{u})$ from the syntactic subspace $\mathcal{S}$ and reconstruct the AST of the code snippet (Sect.~\ref{probe}). In the following subsections, we go through these processes in detail, explain the nature of the vectors $(\pmb{d}, \pmb{c}, \pmb{u})$ and how they can be used to define a probe that recovers the whole AST from a hidden representation space of a model.

\begin{figure}[!ht]
	\centerline{\includegraphics[width=\linewidth]{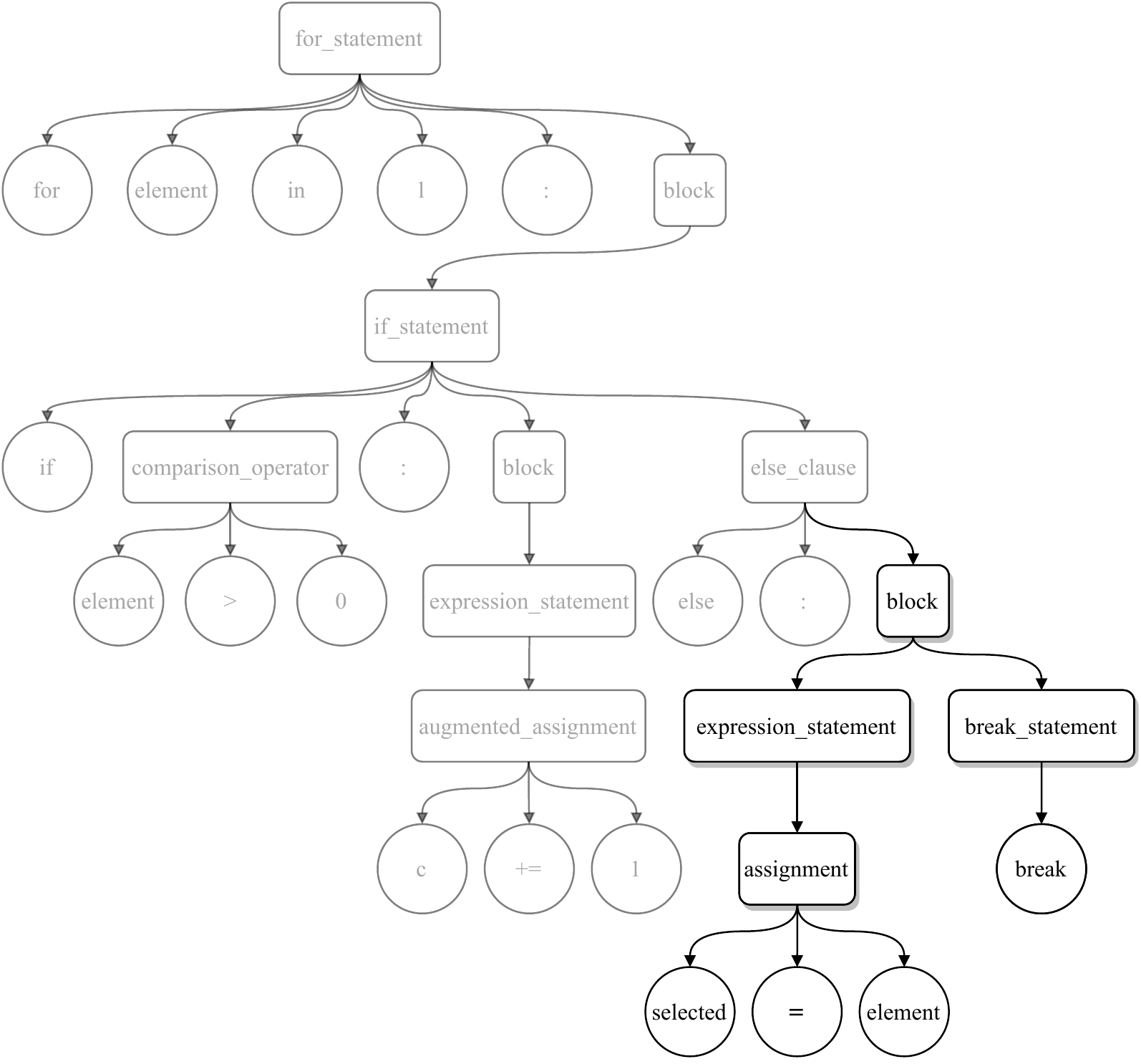}}
	\caption{AST of our running example. The rounded rectangles are non-terminal nodes whereas the circles are terminal nodes.}
	\label{fig:ast-running-example}
\end{figure}

As a running example let us consider the following Python code snippet. Its associated AST is depicted in Fig.~\ref{fig:ast-running-example} and we use it to illustrate our approach.
\begin{lstlisting}[style=mypython,caption={Running example.}]
for element in l:
    if element > 0:
        c+=1
    else:
        selected = element
        break
\end{lstlisting}

\subsection{AST to binary tree}
\label{ast2binary}

\begin{figure}[!t]
	\centerline{\includegraphics[width=\linewidth]{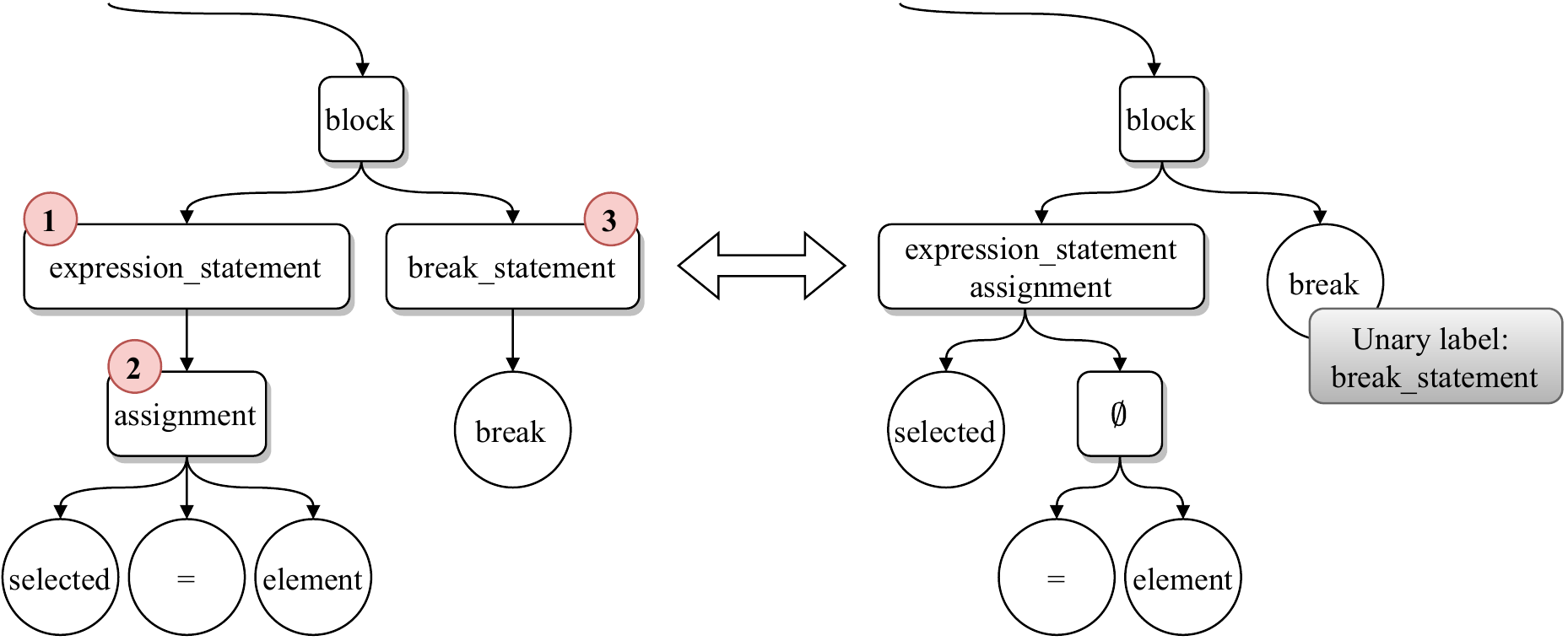}}
	\caption{Excerpt of the binary tree associated to the AST of the running example (Fig.~\ref{fig:ast-running-example}).}
	\label{fig:ast-to-binary}
\end{figure}

First, to transform an abstract syntax tree into a binary tree, the unary and $n-$ary nodes need to be binarized. Since the binary tree associated to the full AST of our running example would not fit the paper, we plot the binary tree that corresponds to the \textit{else block} in the non-shaded part of Fig.~\ref{fig:ast-to-binary}.

Given an AST, we use the following rules taken from~\cite{shen2018straight} to convert it into a \textit{binary parse tree}:

\begin{itemize}
    \item If a non-terminal node is unary, it is merged with its child. An example of the application of this rule is shown in node \encircle{1} of Fig.~\ref{fig:ast-to-binary}. In this example, the non-terminal \textit{expression\_statement} is merged with \textit{assignment} to form a new non-terminal node.
    
    \item If a node is $n-$ary, a special node $\emptyset$ is added to binarize it. An example of the application of this rule is shown in node \encircle{2} of Fig.~\ref{fig:ast-to-binary}. In this example, a non-terminal node $\emptyset$ is added to binarize the 3-ary non-terminal \textit{assignment} node.
    
    \item If a non-terminal chain ends with a unique terminal node then the chain is deleted from the binary tree and its label is added to the terminal node. An example of the application of this rule is shown in node \encircle{3} of Fig.~\ref{fig:ast-to-binary}. The non-terminal \textit{break\_statement} is removed and its terminal node is labeled with its label.
\end{itemize}

Given a binary tree, recovering its corresponding AST is relatively straightforward. The process consists of removing the $\emptyset$ nodes, reconnecting the tree and expanding the unary chains.

\subsection{Binary tree to tuple}
\label{binary2tuple}
In this second step, the binary tree is transformed into the vectors tuple $(\pmb{d},\:\pmb{c},\:\pmb{u})$. The vector $\pmb{d}$ encodes the structural information of the AST whereas $\pmb{c}$ and $\pmb{u}$ encode the labeling information.

\noindent\textbf{Encoding vector $\pmb{d}$}. To construct $\pmb{d}$, we follow the definition of~\cite{shen2018straight}.

\begin{definition}
\label{def:syntacticdistance}
Given a binary parse tree whose leaves are the words $w_0,\dots,w_n$. The syntactic distances of this tree is a vector of scalars $\pmb{d}=(d_1,\dots,d_{n})$ such that
$$\text{sign}(d_i - d_j) = \text{sign}(\tilde{d}^{i-1}_i - \tilde{d}^{j-1}_j)\text{ for all }0\leq i,j\leq n,$$
where $\tilde{d}^{j}_i$ is the height of the lowest common ancestor for two leaves $(w_i,w_j)$. This means that a vector $\pmb{d}$ corresponds to the syntactic distances of a given tree if and only if it induces the same ranking order as $(\tilde{d}^{0}_1,\dots,\tilde{d}^{n-1}_n)$~\footnote{Note that the vector $(\tilde{d}^{0}_1,\dots,\tilde{d}^{n-1}_n)$ also defines valid syntactic distances.}.
\end{definition}
\noindent\textbf{Encoding vectors $\pmb{c}$ and $\pmb{u}$}. Obtaining the vectors $\pmb{c}$ and $\pmb{u}$ is more straightforward. In particular, $\pmb{c}$ is a vector of the same dimension as $\pmb{d}$ that contains the label of the lowest common ancestor for every two consecutive tokens. Whereas, $\pmb{u}$ is a vector whose dimension is the number of terminals containing the unary labels.

For sake of completeness, we replicate here the algorithms of~\cite{shen2018straight} performing the required adaptations. The procedure to convert a binarized AST into a tuple is shown in Algorithm~\ref{alg:ast2tuple}. Whereas the Algorithm~\ref{alg:tuple2ast} describes how to recover the AST from the tuple.

For instance, let us consider the excerpt in Fig.~\ref{fig:ast-to-binary} (right). The tree has four leaves thus the size of the vectors $\pmb{d}$, $\pmb{c}$ and $\pmb{u}$ are three, three and four respectively. In particular, the vector $(2,1,3)$ corresponds to $(\tilde{d}^{0}_1,\:\tilde{d}^{1}_2,\:\tilde{d}^{2}_3)$. Hence, every vector $\pmb{d}$ that verifies the ranking $d_3 > d_1 > d_2$ is valid. On the other hand, the vectors $\pmb{c}$ and $\pmb{u}$ are $($expresion\_statement-assignment, $\emptyset,\text{ block})$ and $(\emptyset$, $\emptyset$, $\emptyset$, break\_statement$)$ respectively. In the context of $\pmb{u}$, the symbol $\emptyset$ means that the terminal node does not have an unary label.

\begin{algorithm}
	\caption{Binary tree to $(\pmb{d},\pmb{c},\pmb{u})$ function extracted from~\cite{shen2018straight}}\label{alg:ast2tuple}
	\begin{algorithmic}[1]
		\Function{tree2tuple}{node}
		\If{node is leaf}
		    \State $\pmb{d}\:\gets\:[]$
		    \State $\pmb{c}\:\gets\:[]$
		    \State $h\:\gets\:0$
		    \If{node has unary\_label}
		        \State $\pmb{u}\:\gets\:[\text{node.unary\_label}]$
		    \Else
		        \State $\pmb{u}\:\gets\:[\emptyset]$
		    \EndIf
		 \Else
		    \State $l,r\:\gets\:\text{children of node}$
		    \State $\pmb{d}_l,\:\pmb{c}_l,\:\pmb{u}_l,\:h_l\:\gets \textsc{tree2tuple}(l)$
		    \State $\pmb{d}_r,\:\pmb{c}_r,\:\pmb{u}_r,\:h_r\:\gets\:\textsc{tree2tuple}(r)$
		    \State $h\:\gets\:\max(h_l,h_r) + 1$
		    \State $\pmb{d}\:\gets\:\pmb{d}_l\:++\:[h]\:++\:\pmb{d}_r$ \Comment{The operator ++ means \textit{concat}}
		    \State $\pmb{c}\:\gets\:\pmb{c}_l\:++\:[\text{node.c\_label}]\:++\:\pmb{d}_r$
		    \State $\pmb{u}\:\gets\:\pmb{u}_l\:++\:\pmb{u}_r$
		 \EndIf
	
    \Return $\pmb{d},\:\pmb{c},\:\pmb{u},\:h$
	\EndFunction
	\end{algorithmic}
\end{algorithm}

\begin{algorithm}
	\caption{$(\pmb{d},\pmb{c},\pmb{u})$ to binary tree function extracted from~\cite{shen2018straight}} \label{alg:tuple2ast}
	\begin{algorithmic}[1]
		\Function{tuple2tree}{$\pmb{d},\pmb{c},\pmb{u}$}
		\If{$\pmb{d} == []$}
		    \State node\:$\gets$\:Leaf$(\pmb{u}[0])$
		 \Else
		    \State $i\:\gets\:\text{argmax}_i(\pmb{d})$
		    \State child$_l\:\gets\:\textsc{tuple2tree}(\pmb{d}_{<i},\:\pmb{c}_{<i},\:\pmb{u}_{\leq i})$
		    \State child$_r\:\gets\:\textsc{tuple2tree}(\pmb{d}_{>i},\:\pmb{c}_{>i},\:\pmb{u}_{> i})$
		    \State node\:$\gets$\:Node$(\text{child}_l,\:\text{child}_d,\:\pmb{c}[i])$
		 \EndIf
		 
    \Return node
	\EndFunction
	\end{algorithmic}
\end{algorithm}

\begin{figure}[!t]
    \centering
    \includegraphics[width=.9\linewidth]{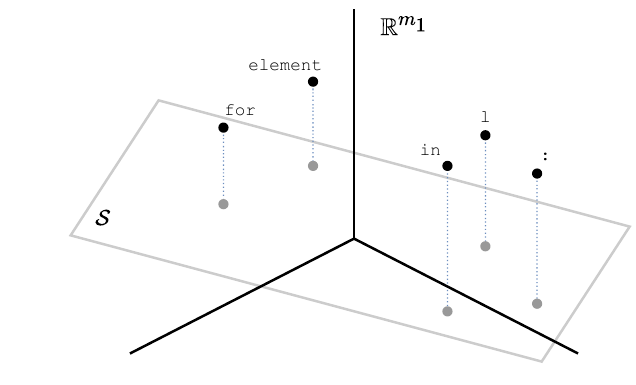}
    \caption{Visualization of the projection. The dotted blue lines represent the projection $P_\mathcal{S}$ of the token representations onto the syntactic subspace.}
    \label{fig:projection}
\end{figure}

\begin{figure}[!t]
    \centering
    \includegraphics[width=.85\linewidth]{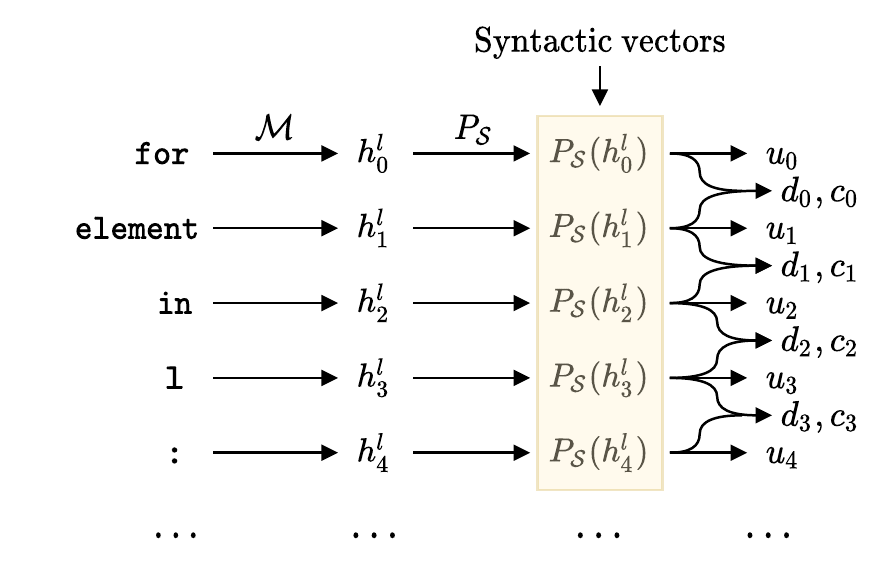}
    \caption{Overview of the AST-Probe. The syntactic vectors are obtained using the projection $P_\mathcal{S}$ (see Fig.~\ref{fig:projection}).}
    \label{fig:probe_overview}
\end{figure}

\subsection{Syntactic subspace and probing classifier}
\label{probe}
Let us denote $\mathcal{M}$ a deep model that receives a sequence $l$ of code tokens $w_0^l,\dots,w_n^l$ as input and outputs a vector representation for each token $h_0^l,\dots,h_n^l \in \mathbb{R}^{m_1}$. Additionally, let us assume that the tuple $(\pmb{d},\:\pmb{c},\:\pmb{u})$ is a valid tuple associated to the AST of the code sequence $l$.

Our hypothesis is that there exists a \textit{syntactic subspace} $\mathcal{S} \subseteq \mathbb{R}^{m_1}$ that encodes the AST information (Fig.~\ref{fig:projection}). The objective of the AST-Probe is to learn a syntactic subspace $\mathcal{S}$ from a hidden representation space of the model $\mathcal{M}$ and assess how well the tuple $(\pmb{d},\:\pmb{c},\:\pmb{u})$ can be predicted from this subspace. We learn $\mathcal{S}$ using our probe by projecting the representations $h_i^l$ onto $\mathcal{S}$. These projected representations are then used to predict the vectors $\pmb{d}$, $\pmb{c}$ and $\pmb{u}$ using the geometry inside $\mathcal{S}$, \textit{i.e.}, the dot product.

Knowing the subspace $\mathcal{S}$, each word vector can be decomposed into a sum of two projections:
$$
h_i^l =P_\mathcal{S}(h_i^l) + P_{\mathcal{S}^ \perp}(h_i^l),
$$
$P_\mathcal{S}(h_i^l)$ contains the syntactic information of $h_i^l$, whereas $P_{\mathcal{S}^ \perp}(h_i^l)$ contains the rest of the original word embedding information. The projection $P_\mathcal{S}(h_i^l)$ is what we call the \textit{syntactic vector} of $w_i^l$.

An overview of the probe is shown in Fig.~\ref{fig:probe_overview}. First, we project the word vectors onto $\mathcal{S}$ to get the syntactic vectors. Then, they are used to infer $\pmb{d}$, $\pmb{c}$ and $\pmb{u}$ using the square Euclidean distance, a set of vectors $\mathcal{C}$ and a set of vectors $\mathcal{U}$.

In the remaining of this section, we go through our probe in more detail. Let us consider $B$ a matrix of dimension $m_2\times m_1$ whose $m_2$ rows define an orthonormal basis of $\mathcal{S}$. The vector $Bh_i^l$ corresponds to the coordinates of $P_{\mathcal{S}}(h_i^l)$ with respect to this basis. In the training procedure, we learn $B$ and two sets of vectors $\mathcal{C},\mathcal{U} \subset \mathcal{S}$ forcing the following conditions:
\begin{enumerate}
    \item $\hat d_i := d\left(P_{\mathcal{S}}(h_{i-1}^l),P_{\mathcal{S}}(h_{i}^l)\right)^2=\|Bh^l_{i-1} - Bh^l_{i} \|^2_2$ is a syntactic distance (see Def. \ref{def:syntacticdistance}) for all $1\leq i \leq n$.
    
    \item For all $i=1,\dots,n$, $P_{\mathcal{S}}(h_{i-1}^l) - P_{\mathcal{S}}(h_{i}^l)$ is similar to the vector $v_{c_i}\in \mathcal{C}$ \textit{i.e.}, $\langle P_{\mathcal{S}}(h_{i-1}^l) - P_{\mathcal{S}}(h_{i}^l), v_{c_i} \rangle$ is high.
    
    \item For all $i=0,\dots,n$, $Bh^l_i$ is similar to the vector $v_{u_i} \in \mathcal{U}$ \textit{i.e.}, $\langle P_{\mathcal{S}}(h_i^l), v_{u_i} \rangle$ is high.
\end{enumerate}

$|\mathcal{C}|$ and $|\mathcal{U}|$ are the number of distinct labels that appear in the vectors $\pmb{c}$ and $\pmb{u}$, respectively. Since $\mathcal{C},\mathcal{U} \subset \mathcal{S}$, we define these vectors with respect to the orthonormal basis induced by $B$. Therefore, the number of total parameters in the probe is $m_1\cdot m_2 + m_2 (|\mathcal{C}| + |\mathcal{U}|)\in \mathcal{O}(m_2)$. It is important to note that the complexity of the probe is given by the dimension of $\mathcal{S}$ \textit{i.e.,} $m_2$. Thus, our probe fits the definition of a simple probing classifier. 



In order to achieve optimality of the subspace $\mathcal{S}$, we minimize a loss function composed of the sum of three losses and a regularization term: $\mathcal{L}=\mathcal{L}_{\pmb{d}} + \mathcal{L}_{\pmb{c}} + \mathcal{L}_{\pmb{u}} + \lambda\cdot\text{OR}(B)$. The first loss is a pair-wise learning-to-rank loss:
$$
\mathcal{L}_{\pmb{d}} = \sum_{i,j>i} \text{ReLU}\left(1 - \text{sign}(\tilde{d}_i^{i-1} - \tilde{d}_j^{j-1})(\hat{d}_i - \hat{d}_j)\right).
$$
We use this loss since we only need the ranking information of $\pmb{d}$ to recover the AST~\cite{shen2018straight} \textit{i.e.}, we want $\pmb{d}$ to induce the same ranking as $(\tilde{d}^{0}_1,\dots,\tilde{d}^{n-1}_n)$. The second loss $\mathcal{L}_{\pmb{c}}$ is defined as follows:
$$\mathcal{L}_{\pmb{c}} = - \sum_{i} \log \frac{\exp{\left(\left\langle P_{\mathcal{S}}(h_i^l) - P_{\mathcal{S}}(h_{i+1}^l), v_{c_i}\right\rangle\right)}}{\sum_{v\in\mathcal{C}}\exp{\left(\left\langle P_{\mathcal{S}}(h_i^l) - P_{\mathcal{S}}(h_{i+1}^l), v\right\rangle\right)}}.$$

Using the cross-entropy loss together with the softmax function, we ensure that the most similar vector to $P_{\mathcal{S}}(h_i^l) - P_{\mathcal{S}}(h_{i+1}^l)$ in the set $\mathcal{C}$ is $v_{c_i}$. $\mathcal{L}_{\pmb{u}}$ is defined similarly:
$$\mathcal{L}_{\pmb{u}} = - \sum_{i} \log \frac{\exp{\left(\langle P_{\mathcal{S}}(h_i^l), v_{u_i}\rangle\right)}}{\sum_{v\in\mathcal{U}}\exp{\left(\langle P_{\mathcal{S}}(h_i^l), v\rangle\right)}}.$$

Finally, the regualization term is defined as follows~\cite{limisiewicz2020introducing}:
$$\text{OR}(B) = \|BB^\text{T}-I\|_\text{F}^2,$$
where $\|\cdot\|_F$ denotes the Frobenius norm and $I$ the identity matrix of dimension $m_2$. We add this component to the loss in order to force the basis, \textit{i.e.}, the rows of $B$, to be orthonormal.




\section{Experiments}
\label{sec:experiments}
In this section, we go through our experimental setup in detail. In particular, we discuss the data and models used in our experiments as well as how we assess the effectiveness of our probe using proper evaluation metrics. To evaluate the relevance of our probe, we articulate our experiments around the following research questions:

-- \textbf{RQ1}: \textit{Can the AST-Probe learn to parse on top of any informative code representation~\footnote{\textit{any informative representation} is a term used in~\cite{hewitt2019structural} to refer to representations that do not contain lots of information and are simple.}?}
    
    We follow the same procedure as in~\cite{hewitt2019structural} to check if the AST-Probe is valid. That is, we try to demonstrate that it does not deeply look for information and just exposes what is already present in the hidden representations~\cite{maudslay2020tale}. To do so, we compare the accuracy of the AST-Probe using pre-trained language models including two baselines: a non-contextualized embedding layer and a randomly initialized language model. If the AST-Probe is valid, a performance gap should lie between the non-baseline models and the baselines. 

-- \textbf{RQ2}: \textit{Which pre-trained language model best encodes the AST in its hidden representations?}
    
    We compare a total of six pre-trained language models for three programming languages and assess which one best encodes the AST of input codes.
    
-- \textbf{RQ3}: \textit{What layers of the pre-trained language models encode the AST better?}
    
    We apply our probe to specific hidden representation spaces of intermediate layers of the models and compare the probe effectiveness over the layers.

-- \textbf{RQ4}: \textit{What is the dimension of the \textit{syntactic subspace} $\mathcal{S}$?}

    To end our experiments, we are interested in how compact is the \textit{syntactic subspace} in the hidden representation spaces of the pre-trained language models.
    
\noindent \textbf{Data availability.} All experimental data and code used in this paper are available at~\url{https://doi.org/10.5281/zenodo.7032076}.

\begin{table*}[!t]
\caption{Summary of the models used in our experiments.}
    \renewcommand{\arraystretch}{1.2}
    \setlength{\arrayrulewidth}{.5pt}
    \centering
    \small
    \begin{tabular}{lcccc} \toprule
        \textsc{Model} & \textsc{Architecture} & \textsc{Number of layers} & \textsc{Training data} & \textsc{Dataset} \\ \midrule
        CodeBERT-0 & embedding layer & 1 & code/doc bimodal & CodeSearchNet~\cite{husain2019codesearchnet} \\
        CodeBERTrand & CodeBERT, init. random & 12 & no training & $-$ \\ \midrule
        CodeBERT~\cite{feng2020codebert} & transformer encoder & 12 & code/doc bimodal & CodeSearchNet~\cite{husain2019codesearchnet} \\ 
        GraphCodeBERT~\cite{guo2020graphcodebert} & transformer encoder & 12 & code/doc bimodal + data flow & CodeSearchNet~\cite{husain2019codesearchnet} \\ 
        CodeT5~\cite{wang2021codet5} & transformer encoder-decoder & 12 & code/doc bimodal & CodeSearchNet~\cite{husain2019codesearchnet} + BigQuery C/C\# \\
        CodeBERTa~\cite{wolf2019huggingface} & transformer encoder & 6 & code/doc bimodal & CodeSearchNet~\cite{husain2019codesearchnet} \\
        RoBERTa~\cite{liu2019roberta} & transformer encoder & 12 & natural language text & BookCorpus~\cite{zhu2015aligning} + English Wikipedia \\
        \bottomrule
    \end{tabular}
    \label{tab:exp-models}
\end{table*}

\subsection{Data}
\label{sec:exp-data}

We choose CodeSearchNet dataset~\cite{husain2019codesearchnet} to conduct our experiments. For all models, we assess their ability to capture the AST of code snippets from three programming languages: Python, Go, and Javascript. For each language, we extract a subset of CodeSearchNet following a ratio of $20000/4000/2000$ samples for training, test, and validation, respectively. 
To extract the AST from the samples, we use the \texttt{tree-sitter} compiler tool~\footnote{\url{https://tree-sitter.github.io/tree-sitter/}}. We follow then the processes described in Sect.~\ref{ast2binary} and Sect.~\ref{binary2tuple} to extract the tuples of vectors $(\pmb{d}, \pmb{c}, \pmb{u})$.

\subsection{Language models}
\label{sec:exp-models}

We compare a broad range of pre-trained language models carefully selected from the state-of-the-art to perform a thorough analysis and bring meaningful discussions to the paper. We summarize the models used in our experiments in Table~\ref{tab:exp-models}. 

The first two rows describe baseline models. CodeBERT-0 refers to an embedding layer initialized with CodeBERT's weights. In this model, the code embeddings are uncontextualized as they are extracted before feeding them to the transformer layers of CodeBERT. In CodeBERTrand, the embedding layer is initialized similarly to CodeBERT-0, but the rest of the layers are randomly initialized. Here, the idea is to randomly contextualize CodeBERT-0. This type of baseline has been used in previous natural language processing related work as strong baselines to evaluate probes~\cite{hewitt2019structural,chi2020finding,conneau2018you}.

Our main models for comparison are CodeBERT, GraphCodeBERT, CodeT5, CodeBERTa and RoBERTa. Note that for CodeT5, we only consider the encoder layers of the encoder-decoder architecture. Finally, RoBERTa is an optimized version of BERT~\cite{devlin2018bert} trained on natural language texts. We include this model in our analysis to check if a model not trained on code is able to capture some understanding of the syntax of programming languages.

\subsection{Evaluation metrics}
\label{sec:exp-metrics}

The output of the AST-Probe is a prediction for the vectors $\pmb{d}$, $\pmb{c}$ and $\pmb{u}$. To get a meaningful interpretation of the effectiveness of the models, we compare the predicted AST, \textit{i.e.}, recovered from the vectors $\pmb{d}$, $\pmb{c}$ and $\pmb{u}$, with the ground-truth AST. To this end, we compute the precision, recall and $F_1-$score over tree constituents~\cite{abney1991procedure}. These three metrics are used in the NLP literature to evaluate constituency parsers and are defined as:
$$\text{Prec} = \frac{|\:\text{Const. in prediction}\:|\cap |\:\text{Const. in ground-truth}\:|}{|\:\text{Const. in prediction}\:|}$$

$$\text{Recall} = \frac{|\:\text{Const. in prediction}\:|\cap |\:\text{Const. in ground-truth}\:|}{|\:\text{Const. in ground-truth}\:|}$$

$$F_1 = \frac{2\cdot \text{Prec}\cdot\text{Recall}}{\text{Prec} + \text{Recall}}$$

In the context of ASTs, we define the constituents as (non-terminal, \textit{scope}) where \textit{scope} indicates the position of the first and last tokens. There is one constituent for each non-terminal node. For instance, the constituents of the AST of Fig.~\ref{fig:ast-to-binary} (left) are the following:

\begin{enumerate}
    \item (block, from \textit{selected} to \textit{break})
    \item (expression\_statement, from \textit{selected} to \textit{element})
    \item (assignment, from \textit{selected} to \textit{element})
    \item (break\_statement, from \textit{break} to \textit{break})
\end{enumerate}

\begin{figure}[!t]
    \centering
    \includegraphics[width=.4\linewidth]{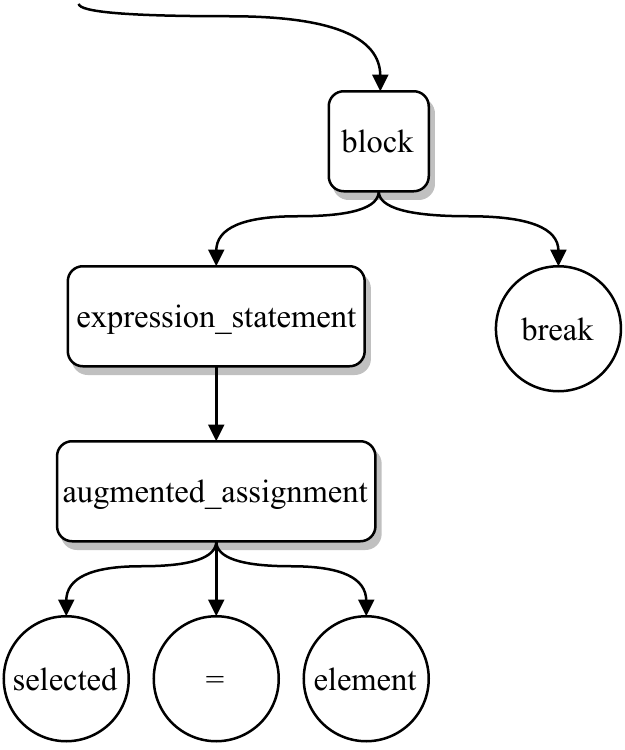}
    \caption{Excerpt of a predicted AST.}
    \label{fig:my_label}
\end{figure}

Let us assume that our predicted AST with respect to that block is the one of the Fig.~\ref{fig:my_label}. Thus, its constituents are the following (\textcolor{dark-green}{\cmark} represents a hit and \textcolor{dark-red}{\xmark} a miss with respect to the true AST):

\begin{enumerate}
    \item (block, from \textit{selected} to \textit{break}) \textcolor{dark-green}{\cmark}
    \item (expression\_statement, from \textit{selected} to \textit{element}) \textcolor{dark-green}{\cmark}
    \item (augmented\_assignment, from \textit{selected} to \textit{element}) \textcolor{dark-red}{\xmark}
\end{enumerate}

In that case, the precision and recall are $2/3$ and $2/4$, respectively.

\subsection{Addressing the research questions}

To answer RQ1/2/3, we train the AST-Probe for each combination of programming language, model and model layer while fixing the dimension of the syntactic subspace to $m_2=128$. And, we report the three evaluation metrics (precision, recall and $F_1-$score). 

To answer RQ4, we select the layer of each model that yields the best $F_1-$score. We then train several configurations of the AST-Probe by varying the maximum number of dimensions of the syntactic subspace starting from eight up until 512 using powers of two. 

\subsection{Training details}

All the considered models were pre-trained using \textit{byte-pair encoding} (BPE)~\cite{sennrich2015neural} which constructs the vocabulary of the model over subwords. Since our analysis is performed over whole-word tokens, we assign to each token representation the average of its subword representations~\cite{hewitt2019structural,wan2022they}. For all models, the dimension of the word embedding space is $m_1=768$. Finally, we set the orthogonal regularization term to a high value $\lambda=5$ in order to ensure orthogonality of the basis induced by the matrix $B$ (see Sect. \ref{probe}).

To perform the optimization, we use the Adam optimizer~\cite{kingma2014adam} with an initial learning rate of $0.001$. After each epoch, we reduce the learning rate by multiplying it by $0.1$ if the validation loss does not decrease. We set the maximum number of epochs to $20$ and use early-stopping with a patience of five epochs. A similar configuration of hyperparameters is used in~\cite{hewitt2019structural} to train the syntax probe.

\section{Results}
\label{sec:results}
In this section, we report the results of our experiments and answer our four research questions. 

\subsection{RQ1 -- Validity of the AST-Probe}

For each model and language, we show in Table~\ref{tab:resultsRQ1} the accuracy of the AST-Probe in the layer of the model that obtained the best $F_1-$score. We can notice a significant performance gap between the baselines and the rest of the models for all considered programming languages and across all metrics. This result validates the probe as it cannot generate ASTs from any informative code representations.

\begin{figure*}[!t]
     \begin{subfigure}[t]{0.45\textwidth}
         \centering
         \includegraphics[width=\textwidth]{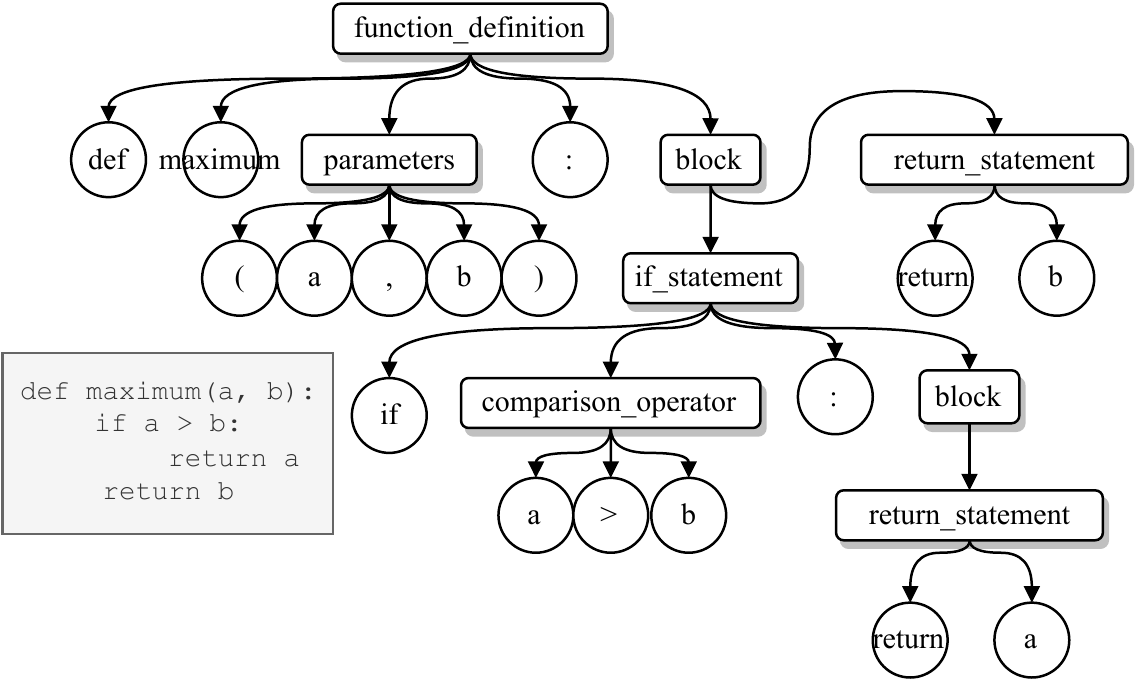}
         \caption{Ground-truth}

     \end{subfigure}
     \begin{subfigure}[t]{0.45\textwidth}
         \centering
         \includegraphics[width=\textwidth]{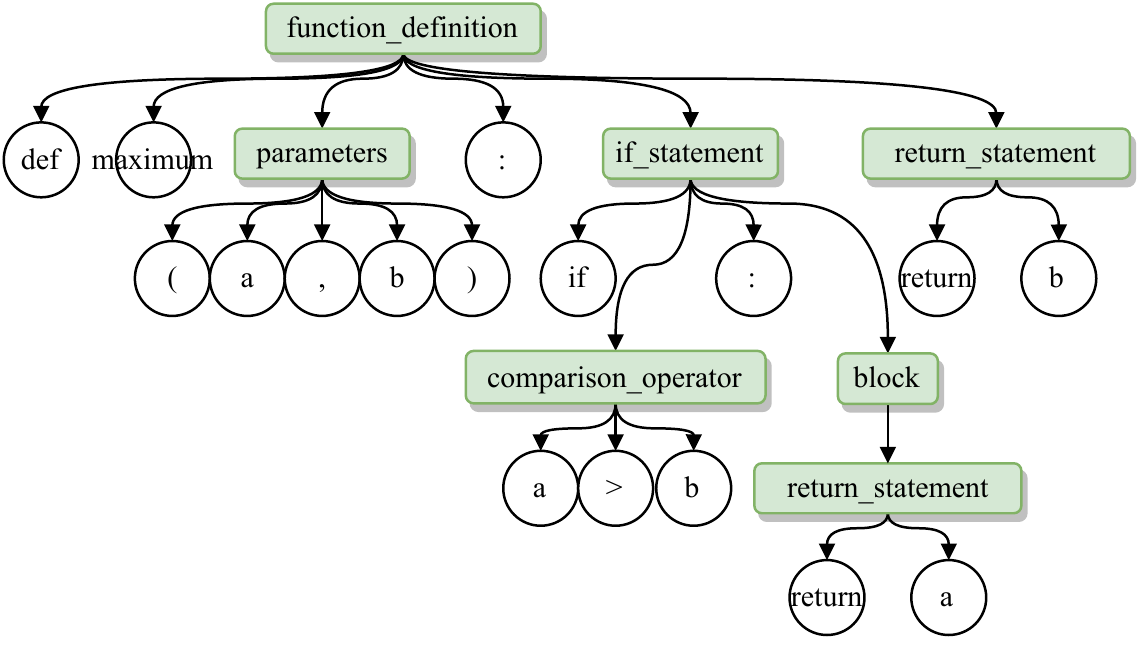}
         \caption{GraphCodeBERT-4}

     \end{subfigure}
     \hfill
     \begin{subfigure}[t]{0.45\textwidth}
         \centering
         \includegraphics[width=\textwidth]{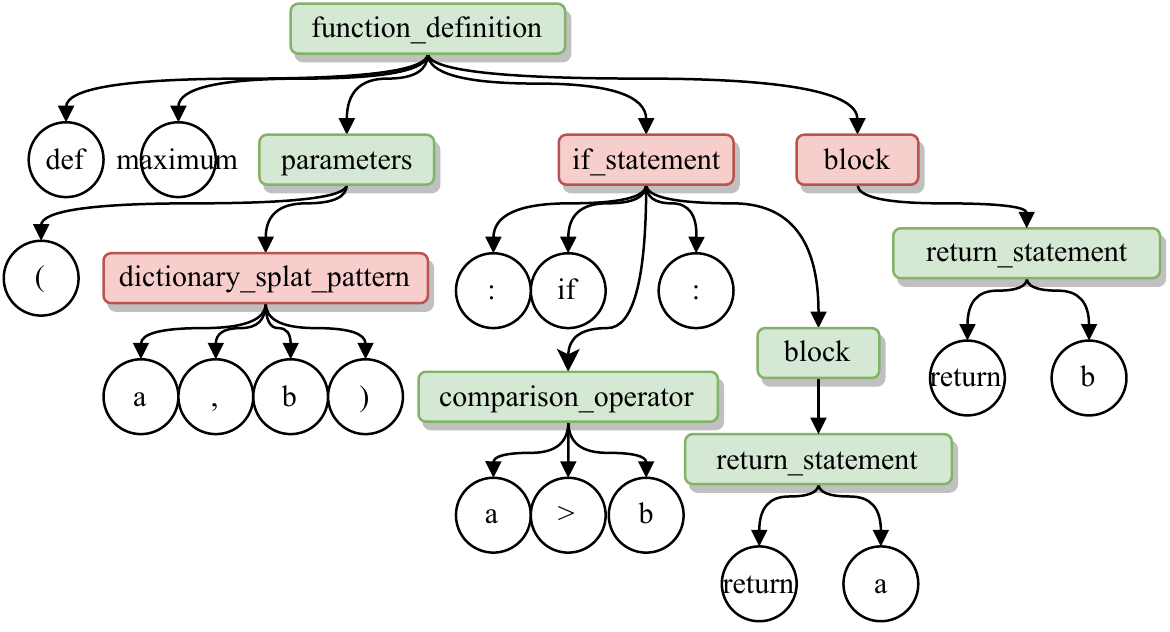}
         \caption{CodeBERTrand-10}

     \end{subfigure}
        \caption{Recovered AST using the probe from the hidden representations of GraphCodeBERT-4 and CodeBERTrand-10. Green rounded rectangles represent constituent hits with respect to the ground-truth AST, and red ones represent misses.}
        \label{fig:astrecovered}
\end{figure*}

Additionally, we report in Fig.~\ref{fig:astrecovered} the AST reconstructed by the probe using the representations of CodeBERTrand-10 (best of the baselines) and GraphCodeBERT-4 (best of the non-baseline models) for a Python code snippet. In this example, GraphCodeBERT performs very well, with a perfect precision score, and the predicted AST is very similar to the ground-truth. As for the baseline, we can observe that the predicted AST contains mistakes even though the chosen code snippet is very simple.

\begin{table}[!ht]
\centering

\begin{subtable}{\linewidth}
\centering
\begin{tabular}{llll}
\hline
                & \multicolumn{3}{c}{\textsc{Metrics}}                                                   \\ \cline{2-4} 
\textsc{Model-BestLayer} & \multicolumn{1}{c}{Precision} & \multicolumn{1}{c}{Recall} & \multicolumn{1}{c}{$F_1$} \\ \hline
CodeBERT-0      & 0.3262                              & 0.4003                            & 0.3573                          \\
CodeBERTrand-10  &   0.3383                            &0.4167                            &  0.3710                         \\ \hline
CodeBERT-5      &  \underline{0.7398}                             & \textbf{0.7657}                           &   \underline{0.7513}                        \\
GraphCodeBERT-4 &  \textbf{0.7468}                             &  \underline{0.7647}                           &  \textbf{0.7545}                         \\
CodeT5-7        &  0.6957                             & 0.7097                           &  0.7016                         \\
CodeBERTa-4     &    0.6620                           & 0.6760                           &    0.6679                       \\
RoBERTa-5       &      0.6724                         & 0.6993                           &    0.6841                       \\ \hline
\end{tabular}
\caption{Python}
\end{subtable}

\begin{subtable}{\linewidth}
\centering 
\begin{tabular}{llll}
\hline
                & \multicolumn{3}{c}{\textsc{Metrics}}                                                   \\ \cline{2-4} 
\textsc{Model-BestLayer} & \multicolumn{1}{c}{Precision} & \multicolumn{1}{c}{Recall} & \multicolumn{1}{c}{$F_1$} \\ \hline
CodeBERT-0      &     0.3358                          & 0.4045                           &   0.3658                        \\
CodeBERTrand-11  &    0.3327                           & 0.4055                           &     0.3642                      \\ \hline
CodeBERT-5      &   \underline{0.7092}                            & \textbf{0.7297}                           &  \underline{0.7186}                         \\
GraphCodeBERT-4 &  \textbf{0.7131}                             & \underline{0.7277}                           &   \textbf{0.7196}                        \\
CodeT5-6        &  0.6650                             & 0.6775                           &  0.6706                         \\
CodeBERTa-5     &    0.6373                           & 0.6561                           &    0.6459                       \\
RoBERTa-8       &  0.6460                              & 0.6724                            &    0.6580                       \\ \hline
\end{tabular}
\caption{JavaScript}
\end{subtable}

\begin{subtable}{\linewidth}
\centering 
\begin{tabular}{llll}
\hline
                & \multicolumn{3}{c}{\textsc{Metrics}}                                                   \\ \cline{2-4} 
\textsc{Model-BestLayer} & \multicolumn{1}{c}{Precision} & \multicolumn{1}{c}{Recall} & \multicolumn{1}{c}{$F_1$} \\ \hline
CodeBERT-0      &     0.4337                          &   0.4932                         &  0.4589                         \\
CodeBERTrand-11  &  0.4403                             &  0.5071                          &    0.4692                       \\ \hline
CodeBERT-5      &  \underline{0.8029}                              & \underline{0.8254}                           &  \underline{0.8134}                         \\
GraphCodeBERT-5 & \textbf{0.8135}                              &  \textbf{0.8314}                          &   \textbf{0.8218}                        \\
CodeT5-8        &    0.7710                           & 0.7889                           &  0.7792                         \\
CodeBERTa-4     &   0.7762                            &   0.7944                         &   0.7846                        \\
RoBERTa-5       &   0.7513                            &  0.7819                          &  0.7653                         \\ \hline
\end{tabular}
\caption{Go}
\end{subtable}
\caption{Results of the AST-Probe for each model and language. We report the best layer of each model together with the baselines. For each metric, the highest value is in bold and the second highest is underlined.} \label{tab:resultsRQ1}
\end{table}


\begin{oframed}
\noindent\underline{\textbf{Answer to RQ1}}: We answer negatively to this research question. That is, the performance gap between the baselines and the other models shows that the AST-Probe is not able to learn to parse on top of any informative code representation. The AST-Probe is thus valid.
\end{oframed}

\begin{figure}[!h]
    \centering
    \includegraphics[width=\linewidth]{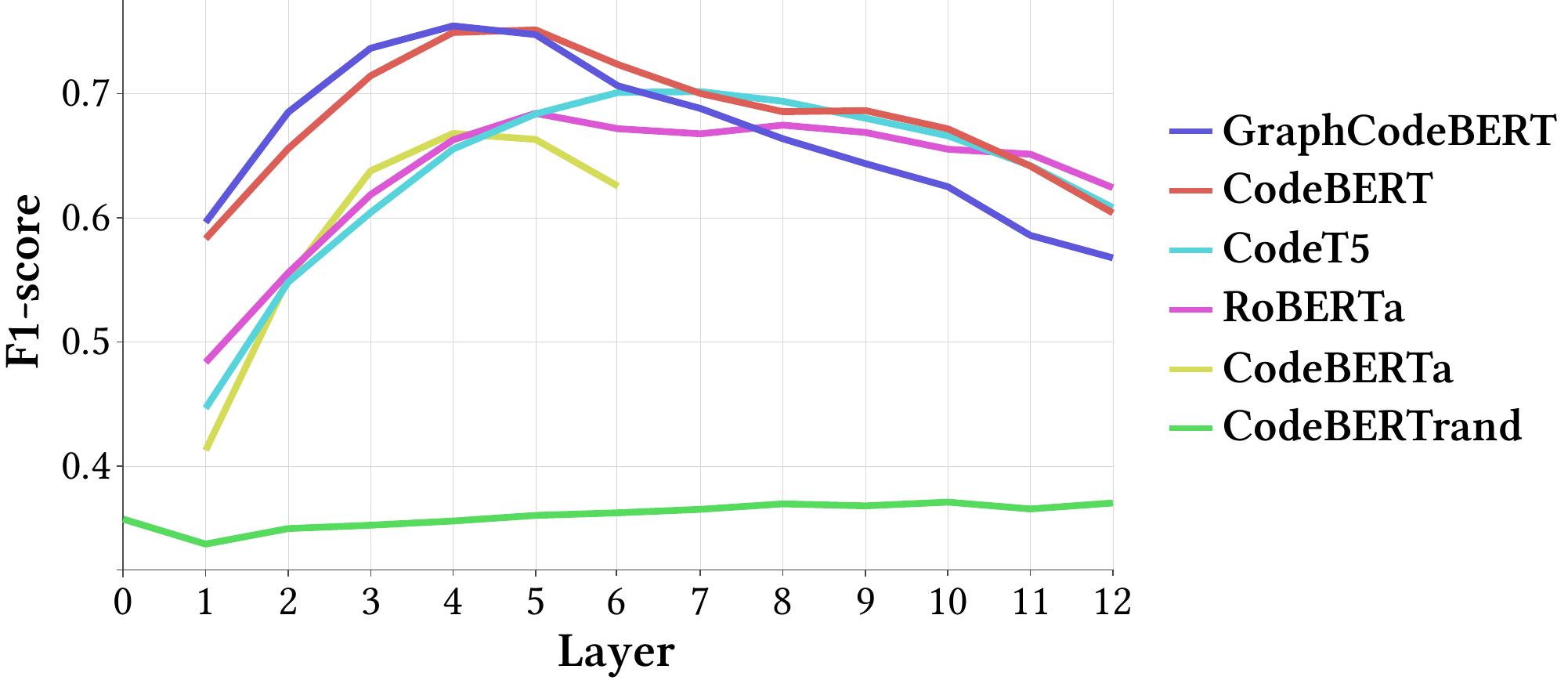}
    \caption{Result of the probe for each model according to their layers. The $x-$axis represents the layer number and the $y-$axis the $F_1-$score. The CodeBERTrand's layer 0 corresponds to CodeBERT-0.}
    \label{fig:layervsf1python}
\end{figure}

\subsection{RQ2 -- Best model}

Overall, the best models are CodeBERT and GraphCodeBERT for all programming languages in terms of $F_1-$score. As the difference in $F_1-$score between both models is very marginal, \textit{e.g.}, $+0.0032\:F_1$ in Python, $+0.0010\:F_1$ in JavaScript, and $+0.0084\:F_1$ in Go in favor of GraphCodeBERT, it is not possible to firmly conclude on which model is better than the other. 

One interesting finding related to this RQ is the fact that, even though trained on text data, RoBERTa competes with models trained on code and is able to understand syntactic information of code. For all languages, RoBERTa achieves similar results with CodeT5. And, in the case of Python and JavaScript, it outperforms CodeBERTa. 


\begin{oframed}
\noindent\underline{\textbf{Answer to RQ2}}: Among all the considered models, both CodeBERT and GraphCodeBERT best capture the AST in their hidden representations for all programming languages.
\end{oframed}

\subsection{RQ3 -- Best layers}

Table~\ref{tab:resultsRQ1} shows that for RoBERTa, the 5th and 8th layers are the best in terms of $F_1-$score. For CodeT5, the best $F_1-$score lies in the 6-7-8th layers. For CodeBERT, GraphCodeBERT, and CodeBERTa, the best $F_1-$score lies in the 4-5th layers. In 
Fig.~\ref{fig:layervsf1python}, we plot the accuracy of the AST-Probe in terms of $F_1-$score for all models with respect to their hidden layer for Python~\footnote{The shapes of the curves are similar for JavaScript and Go.}. For all the models, we can observe a peak in $F_1-$score in the middle layers.


\begin{oframed}
\noindent\underline{\textbf{Answer to RQ3}}: For all models, the AST information is more encoded in their middle layers' representations.
\end{oframed}

\subsection{RQ4 -- Estimating the dimension of $\mathcal{S}$}

\begin{figure}[!t]
    \centering
    \includegraphics[width=\linewidth]{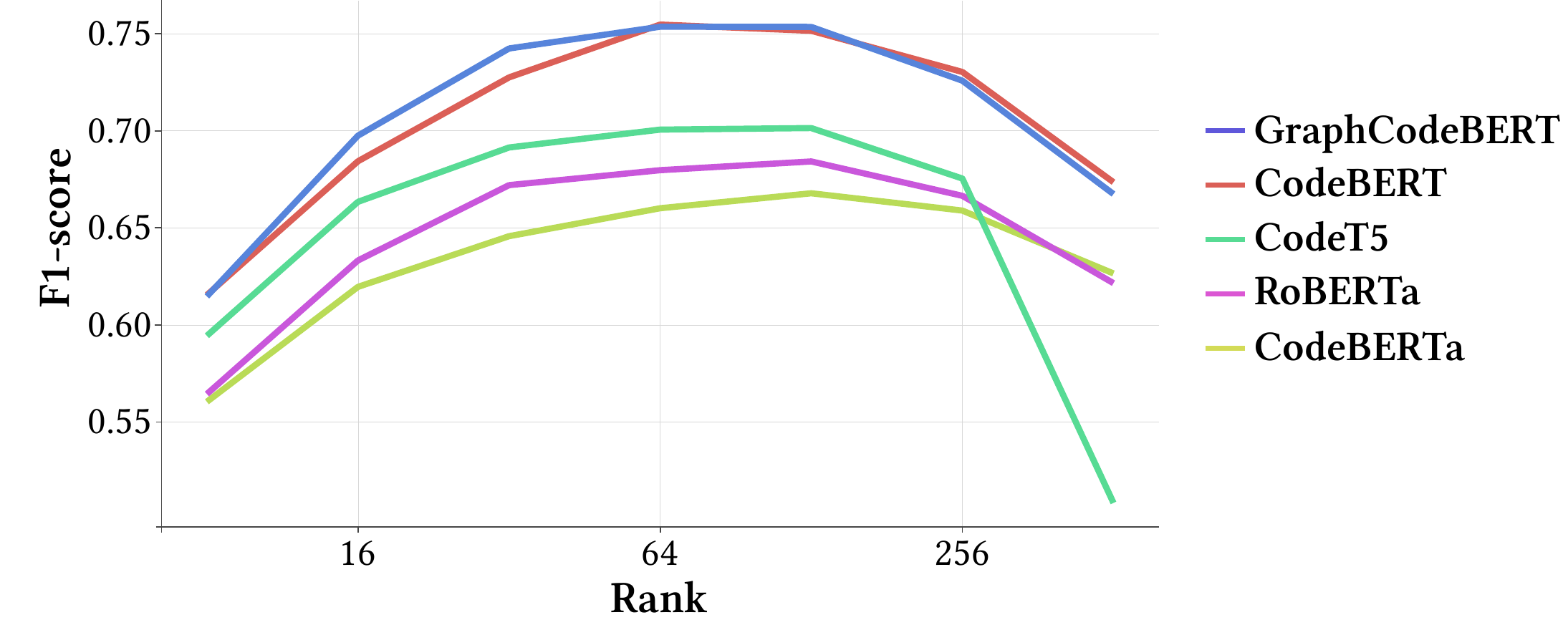}
    \caption{Result of the probe for each model and their best layer according to the dimension of the syntactic subspace for Python. The $x-$axis represents the dimension of the subspace and the $y-$axis the $F_1-$score.}
    \label{fig:layervsrank}
\end{figure}

In Fig.~\ref{fig:layervsrank}, we plot the $F_1-$scores for each model by varying the dimension of the syntactic subspace for Python~\footnote{The shapes of the curves are similar for JavaScript and Go.}. All these curves have a bell shape due to two reasons: \textbf{(1)} when $m_2$ is too small, the number of dimensions is not enough to encode the AST, and \textbf{(2)} when $m_2$ tends to $m_1$, and due to the orthogonality constraint, the projection $P_\mathcal{S}$ tends to produce a rotation of the initial representation space. Thus, if $m_2 \rightarrow m_1$ then $\langle Bx, By \rangle \rightarrow \langle x, y \rangle$ meaning that we are using the full 768-dimensional word vectors, \textit{i.e.}, without preprocessing through a projection, when predicting the AST. It yields a bad performance because the full vectors encode lots of information not only related to syntax. And, we are only interested in the part of the representation space that encodes the syntax, which is ultimately extracted by the orthogonal projection. 

The representation space has 768 dimensions, and the dimension of the syntactic subspace ranges between 64 and 128. It holds across all languages and all models. Only a small part of the representation space is used to encode the AST. It means that the pre-trained language models do not use too many resources to understand the language's syntax as the syntactic information is compactly stored in a relatively low-dimensional subspace.

\begin{oframed}
\noindent\underline{\textbf{Answer to RQ4}}: The dimension of the syntactic subspace ranges between 64 and 128.
\end{oframed}


\section{Discussion}
In this work, we have successfully shown the existence of a \textit{syntactic subspace} in the hidden representations of a set of five pre-trained language models that encode the information related to the code's AST. From these models, GraphCodeBERT and CodeBERT are the ones that best encode this syntactic information of the languages. Furthermore, we show that all the probed models mostly store this information in their middle layers. Finally, our last experiment highlights that the models store the information in a syntactic subspace of their hidden representation spaces with a dimension ranging between 64 to 128.

Our conclusion concerning the best layers is aligned with the results obtained by Troshin and Chirkova~\cite{troshin2022probing} and Wan et. al~\cite{wan2022they}, where they apply different syntactic probes. In NLP, BERT encodes the syntactic knowledge of natural languages in its middle layers~\cite{rogers2020primer}, which lines up with our conclusion for programming languages. An interesting future would be to continue developing probes to search for more properties intricated in the models' representation spaces. Then, try to determine what type of information the models learn in each of their hidden layers.

In their work, Wan et. al~\cite{wan2022they} proposed a structural probe also based on the notion of syntactic distance (see Def.~\ref{def:syntacticdistance}) that can recover unlabeled binary trees from hidden representations and attention layers. Our probe tackles this limitation of their work by recovering full ASTs from hidden representations which is a more complex structure closer to the syntax of the language than unlabeled binary trees. 

Furthermore, we claim that Wan et. al~\cite{wan2022they}'s probe presents two main issues. Firstly, they compare abstract syntax trees (ground-truth) with unlabeled binary trees (predicted using their probe). The metric they use to compare both trees is difficult to interpret as they are not comparing trees of the same nature. The metric also contains a random component. We believe that our approach and experiments cope with this limitation as we compare ground-truth ASTs with ASTs induced using the AST-Probe. Furthermore, we use a metric widely used in the NLP literature to evaluate constituency parsers. \\
The second issue relates to the fact that they extract the unlabeled binary trees from the whole hidden representation spaces, \textit{i.e.}, they take all their dimensions into account. Our analysis of the dimension of the \textit{syntactic subspace} shows that few dimensions of the hidden representation spaces are used by the models to encode syntax. In their work, the authors report that unlabeled binary trees extracted from the attention information produce better results than those extracted from hidden representations. We claim that this can be explained by the fact that they extract unlabeled binary trees from the whole hidden representation spaces of the model. This approach may yield poor results as these representation spaces contain lots of information that are not only related to syntax. 

In fact, in our last experiment, we conclude that the syntactic information is stored compactly in a low-dimensional space with a size ranging between 64 and 128. This means that between $8.3\% \:(64/768)$ and $16.7\% \:(128/768)$ of the dimensions of the full representation space are used to store AST-related information of the code. To the best of our knowledge, this is the first work that carries out this type of analysis. This finding raises questions about the nature of the information encoded in the rest of the dimensions which is an exciting direction for future work.

In this work, we report theoretical findings of the hidden representations of pre-trained language models. We envision that these findings can have broader practical implications. For instance, it seems that GraphCodeBERT recovers ASTs marginally better than CodeBERT. This is correlated with the fact that GraphCodeBERT performs better than CodeBERT in several downstream tasks (\textit{e.g.,} code search, code translation, etc)~\cite{guo2020graphcodebert}. Thus, a good research direction could be to highlight correlations between how well a model encodes syntax and its performance on downstream tasks. Such a practical finding could help us understand what pre-trained language models implicitly learn to perform well in practice. We aim to tackle this interesting question in future work. 

In our experimentations, we measure the performance of recovering the full AST through the $F_1$ score. One possible related research direction would be to analyze which parts of the tree the probe fails to predict using metrics over constituents. It would provide hints on where the pre-trained models struggle to capture syntactical information.

Finally, our work presents several limitations that we plan to tackle in the future. First, we only consider pre-trained language models based on the transformer architecture. Nonetheless, our probe is agnostic of the architecture of the models and programming languages. It can be easily adapted to any type of model involving representation spaces. Then, we choose three common programming languages to analyze. The capacity of our probe to cover a programming language depends on the possibility of extracting the vectors $\pmb{d}, \pmb{c}, \pmb{u}$ from the input code. However, the probe is also language agnostic as it only requires a definition of the language's grammar. 

\label{sec:discussion}

\section{Conclusion and future work}
In this work, we presented a novel probing method, the \textit{AST-Probe}, that recovers whole ASTs from hidden representation spaces of pre-trained language models. The AST-Probe learns a \textit{syntactic subspace} $\mathcal{S}$ that encodes AST-related syntactic information of the code. Then, using the geometry of $\mathcal{S}$, we show how the AST-Probe reconstructs whole ASTs of input code snippets. We apply the AST-Probe to several models and show that the syntactic subspace exists in five pre-trained language models. In addition, we show that these models mostly encode this information in their middle layers. Finally, we estimated the compactness of the syntactic subspace and concluded that pre-trained language models use few dimensions of their hidden representation spaces to encode AST-related information.

In future work, we plan to extend our experiments by including more programming languages. We also want to investigate more pre-trained language models, and different neural network architectures. Furthermore, we plan to study whether a correlation exists between how well a model understands the programming languages' syntax and its performance on code-related tasks. Finally, we intend to compare syntactic subspaces before and after fine-tuning pre-trained language models, for instance, to assess if the models forget about the syntax.

\label{sec:conclusion}


\begin{acks}
José Antonio Hernández López is supported by a FPU grant funded by the Universidad de Murcia. Martin Weyssow is supported by a Fonds de recherche du Qu\'ebec - Nature et technologies (FRQNT) PBEEE scholarship (award \#320621). We thank Dr. Istvan David and Dr. Bentley Oakes for their valuable feedback and comments that helped greatly improve the manuscript.
\end{acks}






\bibliographystyle{ACM-Reference-Format}
\bibliography{sample-base}

\appendix

\end{document}